\documentclass[review]{elsarticle}

\usepackage{hyperref}
\journal{Journal of Visual Communication and Image Representation}

\usepackage{amssymb}
\usepackage{latexsym}

\usepackage{url}
\usepackage{xcolor}
\usepackage{amsmath,amssymb,amsfonts}
\usepackage{hyperref}

\usepackage{algorithm}
\usepackage{algpseudocode}

\usepackage{tabularx}
\usepackage{threeparttable}
\usepackage{booktabs}

\makeatletter
\renewcommand\paragraph{\@startsection{paragraph}{4}{\z@}%
  {-3.25ex \@plus -1ex \@minus -0.2ex}%
  {1.5ex \@plus 0.2ex}%
  {\normalfont\normalsize}}
\makeatother

\begin{document}

\begin{frontmatter}

\title{Domain Generalization for Improved Human Activity Recognition in Office Space Videos Using Adaptive Pre-processing}

\author[1]{Partho Ghosh}
\author[1]{Raisa Bentay Hossain}
\author[1]{Mohammad Zunaed}
\author[1,2]{Taufiq Hasan\corref{cor1}}
\cortext[cor1]{Corresponding author: 
  e-mail.: taufiq@bme.buet.ac.bd;}

\address[1]{mHealth lab, Department of Biomedical Engineering, Bangladesh University of Engineering and Technology (BUET), Dhaka 1205, Bangladesh}
\address[2]{Center for Bioengineering Innovation and Design (CBID), Department of Biomedical Engineering, Johns Hopkins University, Baltimore, MD.}

\begin{abstract}
% Automatic video activity recognition is a critical component of various applications, including surveillance, healthcare, robotics, and more. Recognizing human activities from video data, however, becomes a challenging task when the training data and test data come from different domains, each with its unique characteristics. In such scenarios, domain generalization, the ability to adapt to unforeseen domains, becomes imperative. In this paper we address this challenge in the context of office activity recognition with environmental variability. We propose a set of three pre-processing techniques that can be employed with any video encoder to make it robust against varying environments. Our study demonstrates that the MViT \cite{fan2021multiscale}, a leading state-of-the-art video classification model, and other\cite{8578773, tran2019video, wang2018non} many video encoders along with our pre-processing techniques surpasses state-of-the-art domain adaptation methods in this research area, yielding significant performance improvements in accuracy and F1 score on unseen domains than without our proposed approach. These results highlight the adaptability of our approach, which can play a pivotal role in real-world applications where video data is sourced from different environments. The proposed method paves the way for more reliable video activity recognition systems in domains characterized by data heterogeneity.
Automatic video activity recognition is crucial across numerous domains like surveillance, healthcare, and robotics. However, recognizing human activities from video data becomes challenging when training and test data stem from diverse domains. Domain generalization, adapting to unforeseen domains, is thus essential. This paper focuses on office activity recognition amidst environmental variability. We propose three pre-processing techniques applicable to any video encoder, enhancing robustness against environmental variations. Our study showcases the efficacy of MViT, a leading state-of-the-art video classification model, and other video encoders combined with our techniques, outperforming state-of-the-art domain adaptation methods. Our approach significantly boosts accuracy, precision, recall and F1 score on unseen domains, emphasizing its adaptability in real-world scenarios with diverse video data sources. This method lays a foundation for more reliable video activity recognition systems across heterogeneous data domains.
\end{abstract}


% \begin{keyword}
% \texttt{elsarticle.cls}\sep \LaTeX\sep Elsevier \sep template
% \MSC[2010] 00-01\sep  99-00
% \end{keyword}

\end{frontmatter}

% \linenumbers

\section{Introduction}
\label{sec:introduction}

%----------------------------------------------------------------------Introduction-------------------------------------------------------
Human Activity Recognition (HAR) categorizes human actions using data from sources like sensors and camera footage \cite{ref_1,ref_6}. The widespread adoption of HAR in applications like healthcare, surveillance, and smart environments \cite{ref_10,ref_21}, holds significant benefits for enhancing human safety and quality of life \cite{ref_25}. Human activity classification uses sensor and video data, with video being less common due to challenges in data collection, processing, and interpretation. The rise of body-worn cameras has increased video data availability, categorized as third-person (TPV) or first-person (FPV). For both FPV and TPV domains, the major challenge in automatic HAR includes scale and texture variation, low-resolution, motion-blur, illumination changes and context analysis \cite{ref_26, ref_27}. FPV, on the other hand, poses some additional challenges due to a more dynamic background, self-occlusion (e.g., user’s hand occluding the camera viewpoint), scarcity of information, and unstable perspective. These challenges have spurred interest in enhancing FPV processing for HAR across diverse environments.

Various machine learning methods like SVM, KNN, Random Forest, CNNs \cite{ref_28}, LSTM \cite{ref_29}, and transformers \cite{ref_30} have been used to enhance HAR performance, but generalization to new and unseen datasets remains a significant challenge \cite{intro**}. Video data is affected by user characteristics such as body shape, skin color, and environment, leading to performance drops on unseen users. Models like R(2+1)D \cite{8578773}, SlowFast \cite{ref_32}, and MVit \cite{fan2021multiscale} gives satisfactory results when trained with mix datasets, however when deployed in domain specific data their performances drop significantly. This issue of domain shift can be addressed through domain adaptation \cite{DA_intro} and domain generalization \cite{dg_intro}. The key difference between domain adaptation and domain generalization is that domain adaptation assumes that the target domain is known during training, while domain generalization does not \cite{ref_39}. This makes domain generalization more challenging, but also more powerful since it allows models to generalize to completely unseen domains \cite{ref_37}. However, current domain generalization methods for HAR are limited, especially in video-based applications where the dynamic nature of FPV data further complicates the problem. This study aims to develop a domain generalization approach for HAR using First Person Video (FPV) data in office environments.

\begin{figure*}[ht]
    \centering
    %\hspace{-20mm}
    \includegraphics[width=\textwidth]{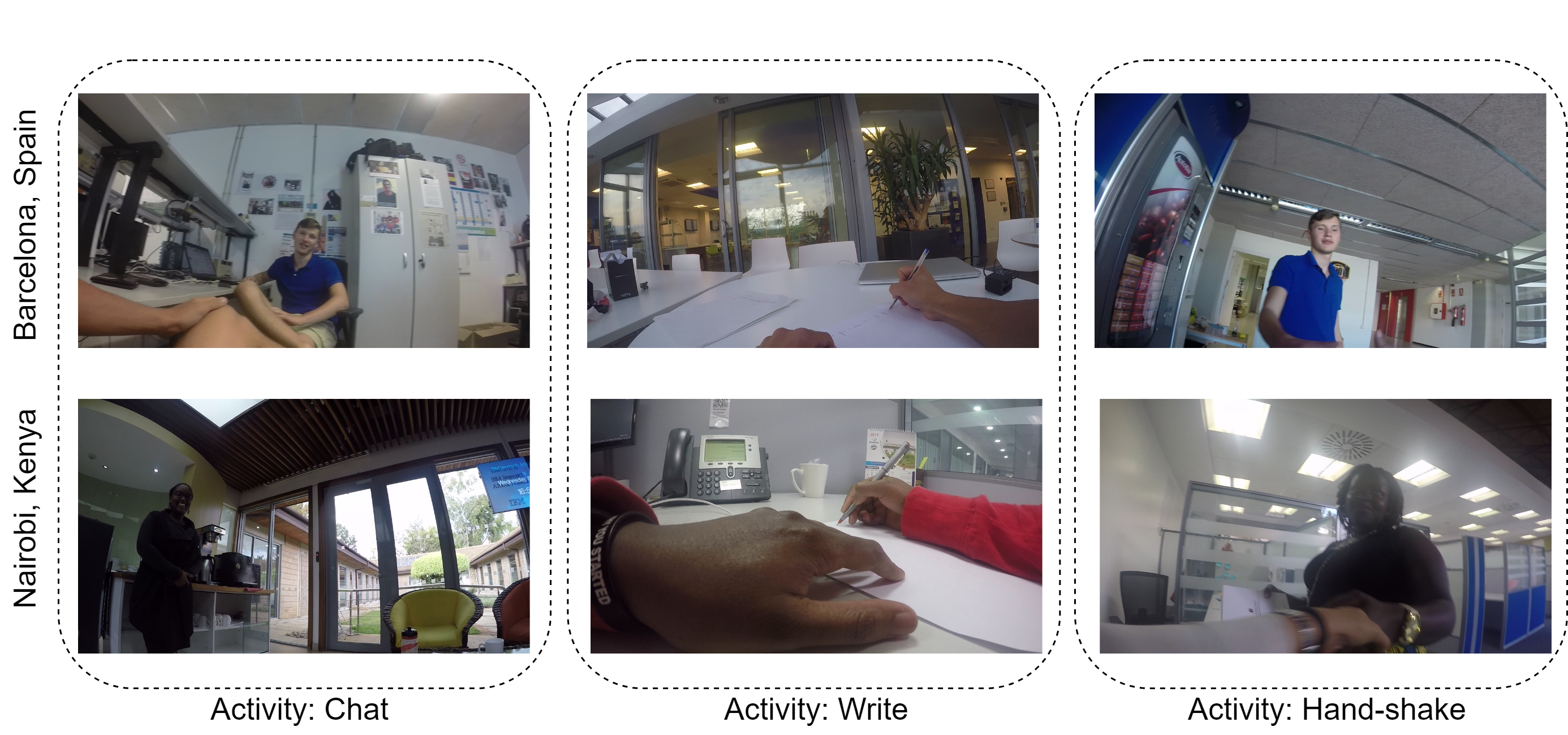}
    %\vspace{-55mm}
    \caption{Differences caused by domain shift. Due to the difference in location, the office environment along with the point of view and person's skin color is completely different even for the same activity category. }
    \label{frames}
\end{figure*}

\begin{figure}[ht]
    \centering
    %\hspace{-20mm}
    \includegraphics[width=0.6\textwidth]{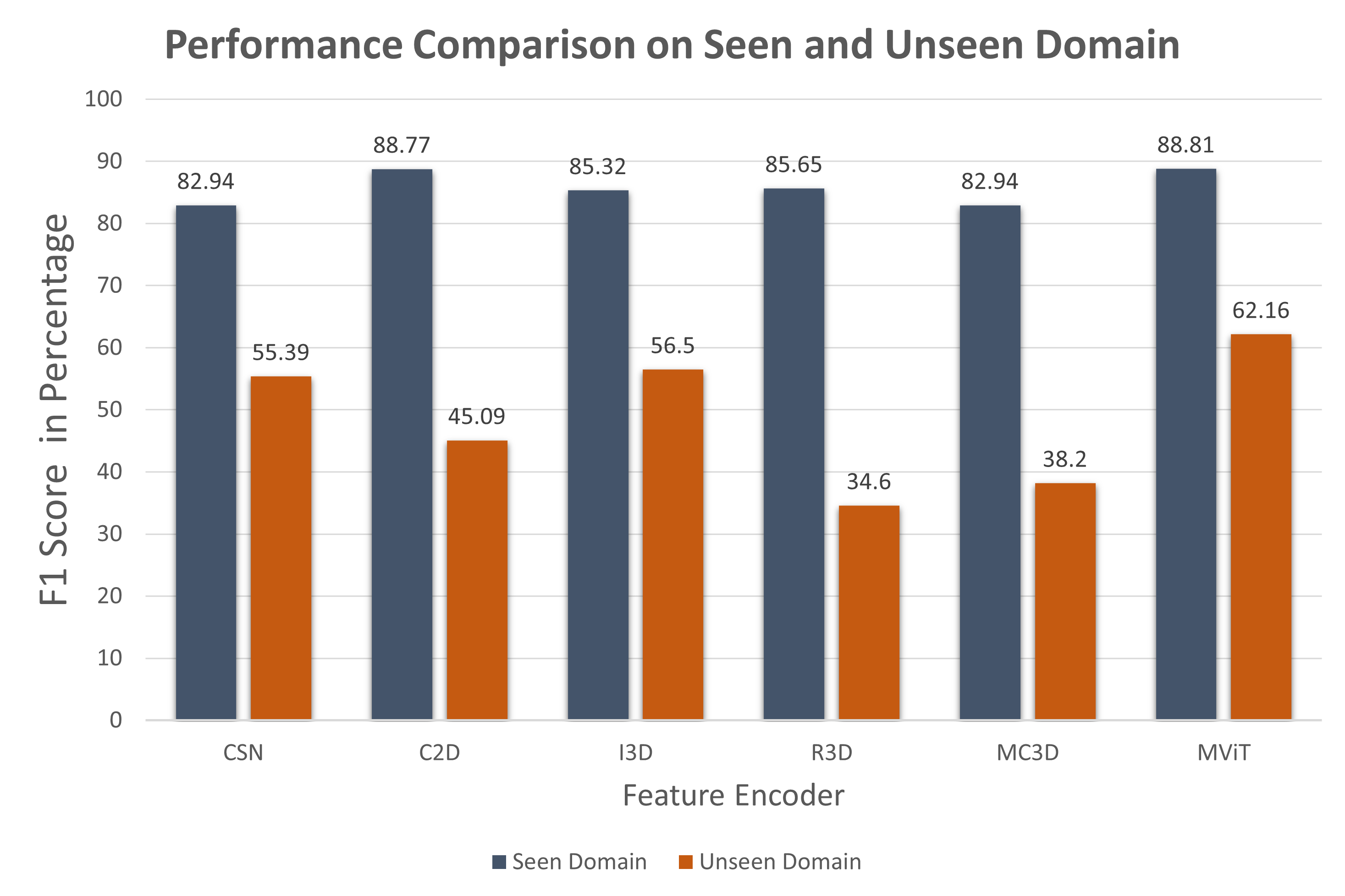}
    %\vspace{-55mm}
    \caption{Evaluation of performance degradation of video classifiers due to domain disparity.}
    \label{chart}
\end{figure}

To grasp the impact of domain shift and the need for domain generalization in human activity recognition, consider the six frames in Fig. \ref{frames}. The first row shows office scenes from Barcelona, Spain, while the second row features Nairobi, Kenya. Each column depicts three activities: chatting, writing, and hand-shaking. Even within the same activity, there are noticeable differences in physical characteristics, skin tones, and office environments. These pronounced dissimilarities between domains reveal a critical challenge: a model trained solely on data from one domain struggles with data from an `unseen' domain due to significant differences, leading to poor activity recognition in new contexts. To further support this, Fig. \ref{chart} compares the performance of various state-of-the-art video feature encoders. These algorithms, trained on data from Barcelona and Oxford (seen domains), show significant performance drops when tested on data from Nairobi (unseen domain). The results highlight the adverse impact of domain shift and how models trained in one environment struggle to generalize to different contexts. This issue motivates our study, which aims to develop a pre-processing framework for better domain generalization in human activity recognition.

In this study, we present a pre-processing techniques to enhance off-the-shelf video classifier performance across various domains. We address the challenges posed by environmental and subject changes in human activity recognition. The main contributions of this paper are five-fold:

\begin{enumerate} 
    \item We introduce a single-layer MLP network with learnable decision parameter embeddings to dynamically determine whether to use raw frames or masked frames for forward propagation, thereby optimizing video feature encoding and improving overall model performance. 
    \item We propose a frame-wise attention mechanism to assign varying degrees of importance to each frame within a video, enhancing the understanding of significant features and refining the performance of the human activity recognition system. 
    \item We design pre-processing techniques, including the adaptive decision-making MLP network and frame-wise attention mechanism, to seamlessly integrate with any video classifier network in a plug-and-play manner, supporting end-to-end training for flexibility and applicability across diverse video classification tasks. 
    \item We employ a single-stream network that utilizes both masked frames and raw frames for classifier training, contrasting with conventional approaches that often use multiple streams \cite{lit_5, carreira2017quo, 8578773}. 
    \item We address the gap in research by applying domain generalization techniques specifically to video-based human activity recognition, which has not been extensively explored compared to other domains such as audio or sensor data. 
\end{enumerate}

The paper is organized as follows: Section 2 reviews related work in HAR using FPVs and Domain Generalization. Section 3 outlines our methodology, with experimental details and results in Section 4. We conclude with a discussion in Section 5 and a final conclusion in Section 6.

% ---------------------------------------------------Figure being added to introduction---------------------------------------------

% --------------------------------------------------------------Table for related work-------------------------------------------------------------

% Table for comparing recent works
\begin{table*}[ht]
\caption{\textsc{Recent Action Recognition Methods Focusing on Domain Adaptation}}
\label{related_work}
\small
\begin{tabularx}{\textwidth}{>{\hsize=0.4\hsize}X|>{\hsize=0.1\hsize}X|X|>{\hsize=0.5\hsize}X}
\hline
\textbf{Title of Work} & \textbf{Year} & \textbf{Methodology} & \textbf{Relevance with our work} \\ \hline

Dual-Head Contrastive Domain Adaptation for Video Action Recognition (CO$^{2}$A)\cite{da2022dual} &2022 & Introduced a novel Unsupervised Domain Adaptation (UDA) approach for action recognition in videos, utilizing a two-headed deep architecture that combines cross-entropy and contrastive losses from different network branches. & Used the target domain data and the pseudo-labels in the training pipeline. \\ \hline
Domain Generalization through Audio-Visual Relative Norm Alignment in First Person Action Recognition\cite{planamente2022domain} &2022 & Introduced the first domain generalization approach for egocentric activity recognition, featuring a novel audiovisual loss called Relative Norm Alignment loss, which re-balanced modality contributions across different domains by aligning their feature norm representations. & Addressed environmental bias issue which is relevant to our work. However, used audio data along with video data. \\ \hline
Domain Generalization for Activity Recognition via Adaptive Feature Fusion\cite{qin2022domain} &2022 & Combined domain-invariant and domain-specific representations to enhance model generalization without access to test data during training.  & Directly matches our goal but used sensor data for activity recognition.  \\ \hline
Source-free Video Domain Adaptation by Learning Temporal Consistency for Action Recognition\cite{xu2022source} &2022 & Introduced a novel approach called Attentive Temporal Consistent Network (ATCoN) for Source-Free Video-based Domain Adaptation (SFVDA), which addressed the challenge of adapting video models to different environments without requiring access to source data. & Used target domain data for training as well. \\ \hline
\end{tabularx}
\end{table*}

% ----------------------------------------------------------------------Related work section-----------------------------------------------------

\section{Related Work}

Previous research in action recognition has advanced significantly with transfer learning and diverse deep learning architectures. Early approaches relied on manual feature encoding and traditional machine learning algorithms \cite{lit_1,lit_2}. Techniques like optical flow and Histogram of Oriented Gradients (HOG) were common \cite{lit_3,lit_4}. Neural networks, especially CNNs and RNNs, have automated feature extraction \cite{lit_5,lit_6,lit_8}. Models like C3D and I3D employ 3D convolutions \cite{lit_10,carreira2017quo}. Contrastive learning, pretrained models like MVit \cite{lit_14,fan2021multiscale}, and PyTorch's \emph{PyTorch Video} library \cite{lit_16} have boosted performance.

\begin{figure*}[t]
    \centering
    %\hspace{-20mm}
    \includegraphics[width=0.95\textwidth]{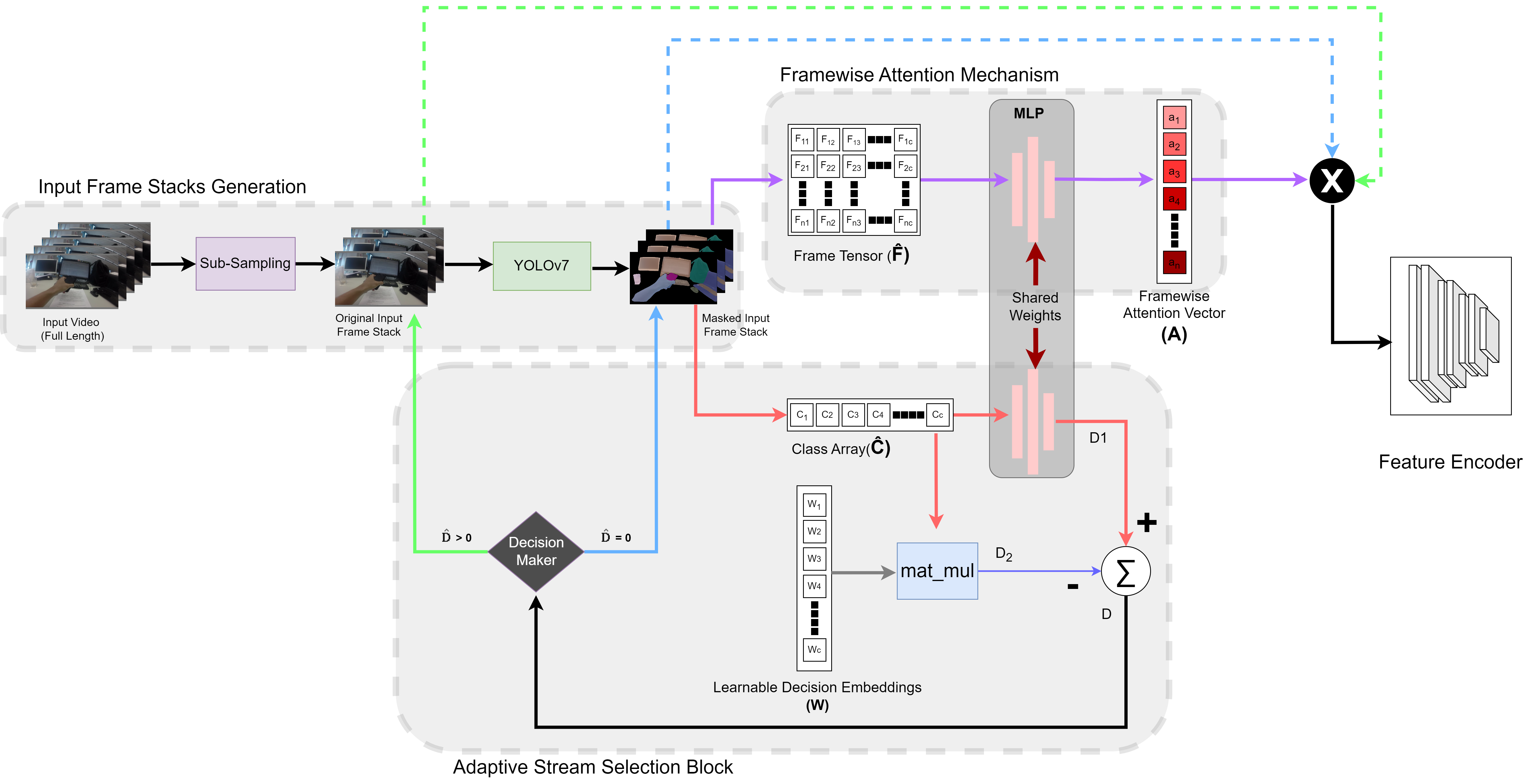}
    %\vspace{-55mm}
    \caption{Proposed Preprocessing Framework. In the input to the framework we are giving the Frame Tensor ($\mathbf{\hat{F}}$), Class Array ($\mathbf{\hat{C}}$), and the Video Segment and at the output we are having the activity class happening in the video segment.}
    \label{network_main}
\end{figure*}

Despite these advancements, domain shift remains a significant challenge, addressed by two primary approaches: domain adaptation and domain generalization. Domain adaptation focuses on improving model performance in a specific target domain by utilizing data from that domain during training. Most of the recent video-based human activity recognition studies addressing domain shift rely on some form of domain adaptation. For instance, Choi \emph{et al.} \cite{dg_5} developed a video domain adaptation method using attention mechanisms, while Zhang \emph{et al.} \cite{dg_6} integrated audio data to enhance adaptation. Techniques like adversarial training and self-supervised learning, such as the Adversarial Correlation Adaptive Network by Xu \emph{et al.} \cite{dg_7} and Video Masked Autoencoders by Tong \emph{et al.} \cite{dg_8}, are also widely used.

In contrast, domain generalization aims to train models that perform well on previously unseen domains without access to data from those domains during training. This approach is particularly valuable in scenarios where acquiring data from all potential target domains is impractical. However, most of the previous works on Domain Generalization were mostly designed for image data. They are primarily divided into feature-based and data-based methods. Feature-based models focus on extracting invariant information shared across domains, whereas data-based methods employ data augmentation. Feature-based methods, like the adversarial autoencoders framework by Li \emph{et al.} \cite{dg_1}, focus on extracting invariant information shared across domains. Data-based methods, as demonstrated by Volpi \emph{et al.} \cite{dg_3}, employ data augmentation to bridge domain gaps, while Bucci \emph{et al.} \cite{dg_4} utilized self-supervised pretext tasks to extract robust representations.

%Despite these advancements, domain shift remains a significant challenge. Domain generalization aims to transfer knowledge from related domains with available training data to previously unseen domains. Previous works mainly focus on image data and are categorized into feature-based and data-based methods. Feature-based models extract invariant information shared across domains, such as Li \emph{et al.} \cite{dg_1} who introduced an adversarial autoencoders framework for learning a generalized latent feature representation. Data-based methods use data augmentation strategies to bridge the domain gap, as demonstrated by Volpi \emph{et al.} \cite{dg_3}, who proposed an iterative procedure to augment the dataset with examples from a fictitious target domain. Self-supervised pretext tasks, as shown by Bucci \emph{et al.} \cite{dg_4}, are also efficient for extracting robust data representations.

%In video-based human activity recognition, domain adaptation methods are commonly used to address domain shift challenges. Choi \emph{et al.} \cite{dg_5} introduced a video domain adaptation method utilizing attention mechanisms and Zhang \emph{et al.} \cite{dg_6} incorporated audio data into their approach. Adversarial training and self-supervised learning techniques have also been utilized, such as the Adversarial Correlation Adaptive Network proposed by Xu \emph{et al.} \cite{dg_7} and Video Masked Autoencoders introduced by Tong \emph{et al.} \cite{dg_8}.

In Table \ref{related_work}, we provide an overview of recent studies in human activity recognition, focusing on their approaches to tackle domain shift, and discuss their relevance to our research. Many of these studies include target domain data in their training, which may not always be practical in real-world scenarios where acquiring data from all potential domains is challenging, highlighting a significant gap in the existing literature.

\begin{figure*}[t]
    \centering
    %\hspace{-20mm}
    \includegraphics[width=0.8\textwidth]{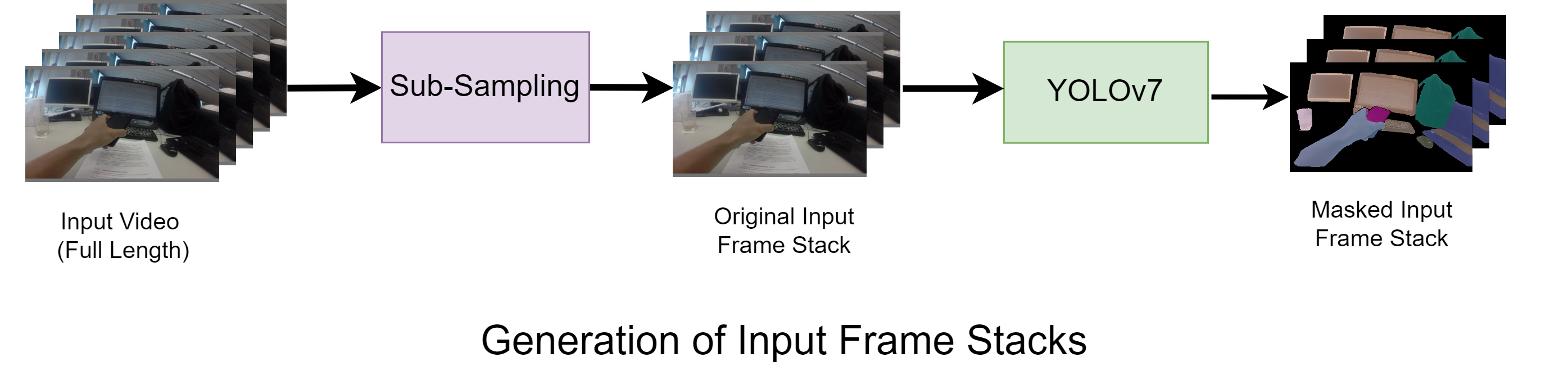}
    %\vspace{-55mm}
    \caption{Frame extracted from video clips are sub-sampled to get `Raw frame stack', which is then passed through YOLLOv7 algorithm to generate `Masked frame stack'.}
    \label{frame_stack}
\end{figure*}

% ----------------------------------------------------------------Methodology-----------------------------------------------------------------------

\section{Method}

In this section, we will provide a brief overview of the problem statement followed by a detailed explanation of our propsed preprocessing framework. The complete workflow of our architecture can be found in Fig. \ref{network_main}.

\subsection{Problem Definition}

In a typical Human Activity Recognition (HAR) problem, we are presented with a training dataset $D^\text{train} = \{(x_i, y_i)\}_{i=1}^{n}$, consisting of features $x \in \mathbb{R}^d$ and corresponding activity categories $y \in \{1, 2, \ldots, C\}$, with dimensions $d$, $n$, and $C$ representing the data, number of samples, and activity categories, respectively. The objective is to build a model $\mathcal{H}: x \rightarrow y$ that accurately recognizes activities both in the training data and in the test dataset.

However, when our dataset contains multiple domains instead of a single domain, we denote the domains in the training set as `source domains' and those in the test set as `target domain'. If we have $\mathcal{K}$ different but related source domains, the training dataset can be expressed as $\mathcal{D}^\text{train} = \{D^1, D^2, \ldots, D^K\}$, where $D^k = \{(x_i^k, y_i^k)\}_{i=1}^{n_k}$ represents the $k$-th source domain with $n_k$ samples. Our objective is to learn a generalized model $\mathcal{H}$ on the $K$ training domains to minimize error on the training dataset and perform well on an `unseen' target domain $\mathcal{D}^\text{test} = \{(x_i, y_i)\}_{i=1}^{n_\text{test}}$, where $n_\text{test}$ is the number of samples in the target domain \cite{qin2022domain}.

Achieving low training error is important, but the model must also perform well on the target domain for effective results. Unlike traditional transfer learning and domain adaptation, which require test domain knowledge, we tackle this challenge using domain generalization methods.

\textbf{Domain Generalization:} Domain generalization involves training a model on a set of source domains to generalize well to unseen target domains by learning features that are invariant to domain-specific variations. Formally, given a set of $m$ source domains $\Delta = \{P^1, \ldots, P^m\}$ and a target domain $P^t \notin \Delta$, with samples $S^d = \{(x_{i}^d, y_{i}^d)\}_{i=1}^{n_d} \sim P^d$ drawn from $m$ source domains, the task is to learn a labeling function $f_{P^t} : \mathcal{X} \rightarrow \mathcal{Y}$ using $S^d$ as training examples \cite{ref_39}.

The following subsections describe different components of the proposed activity classification system.

\subsection{Proposed Framework}

Our proposed preprocessing framework is structured into three primary subsections. The initial subsection focuses on generating input frame stacks, while the subsequent sections are referred to as the ``Adaptive Stream Selection Block" and the ``Framewise Attention Mechanism." 
Each of these subsections are elaborated in the subsequent discussion.

\subsubsection{Generation of Input Frame Stacks}

In the network model shown in Fig. \ref{network_main}, two input frame stacks are used: one from the original video clips and another with masked frames. Each video clip in our dataset is treated as a sequence of image frames. To address the variability in clip lengths, we sub-sample the videos, extracting a fixed number of frames from each. This approach reduces computational load while ensuring the key information from each video is retained. The sub-sampled frames form the first input stack, referred to as the \emph{original input frame stack} or \emph{raw frame stack}. This stack is processed through the YOLOv7 \cite{wang2022yolov7} algorithm to create the second stack, the \emph{masked frame stack}. YOLOv7 performs \textcolor{blue}{semantic segmentation}, assigning a unique color to each object and black to non-object pixels. This masking process highlights regions of interest, enhancing object visibility and making feature detection more efficient and robust. Fig. \ref{frame_stack} illustrates the creation of both the original and masked frame stacks.

However, masking removes important background details like environment, lighting, and object poses, which are essential for activity detection. This loss can harm classifier performance, but including background information risks biasing the algorithm towards the environment instead of the activity. To address this, we include frames from the original video and adaptively switch between the original and masked frame stacks during training. Further details are provided in Sec. \ref{adaptive-stream-selection}.

\begin{figure*}[ht]
    \centering
    %\hspace{-20mm}
    \includegraphics[width=\textwidth]{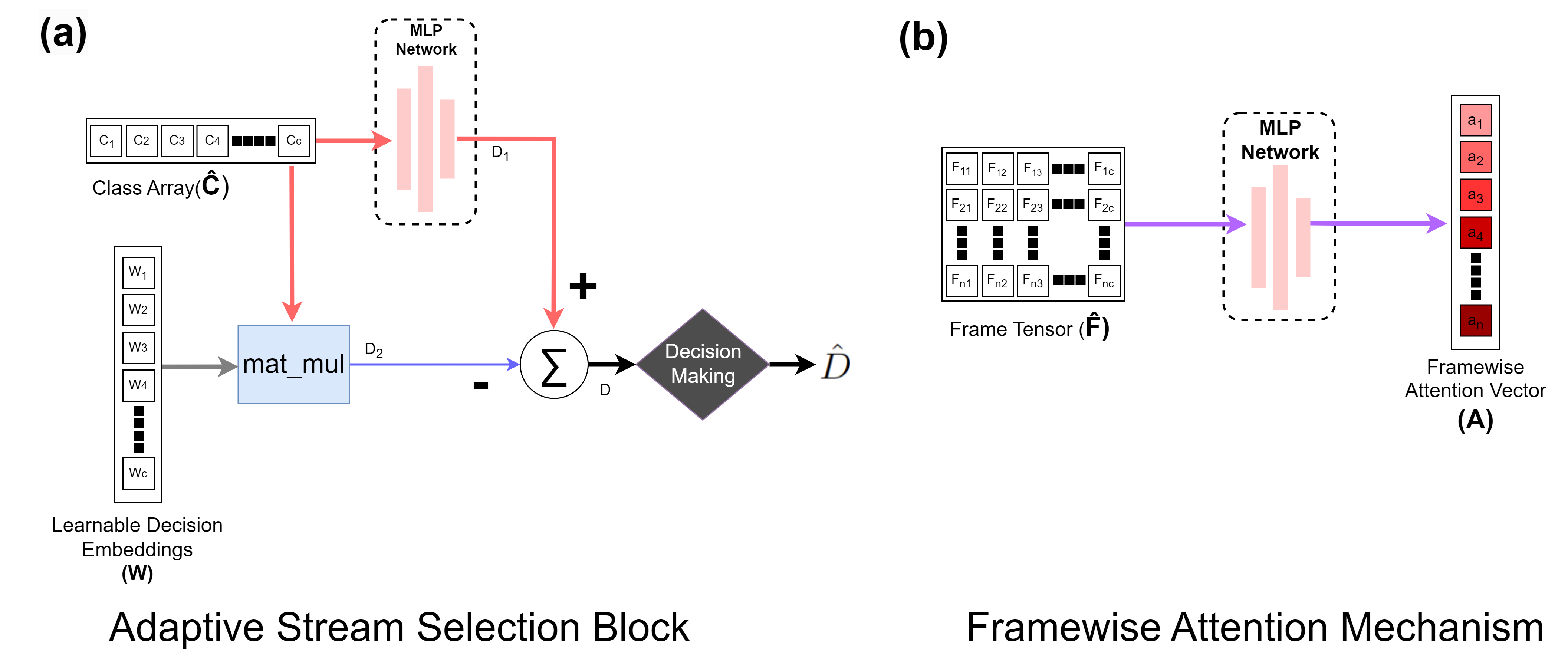}
    %\vspace{-55mm}
    \caption{(a) \textbf{Adaptive stream selection block:} The \emph{c}-dimensional vector class array $\hat{C}$ and the trainable weight vector $W$ are multiplied to generate a scalar $D_2$. Which is subtracted from $D_1$, obtained from the MLP network. The resulting $D$ is processed through a ReLU decision block to produce the output $\hat{D}$, which is either zero or positive. (b) \textbf{Framewise attention mechanism:} The frame tensor $\mathbf{\hat{F}}$ (of size $(n,c)$) is passed through the same MLP network to generate an attention vector $\mathbf{A}$.}
    \label{adaptive_stream}
\end{figure*}

\subsubsection{Adaptive Stream Selection Block}\label{adaptive-stream-selection}

A key focus of our work is to use a single-stream approach instead of the multi-stream networks commonly employed in human activity recognition. The challenge lies in adaptively selecting the appropriate frame stack (original or masked) from two distinct streams. Random selection or relying on a single stream wouldn't solve the domain shift issue. To address this, and recognizing that activity prediction correlates strongly with specific objects in the video, we introduce an `adaptive stream selection block’. Fig. \ref{adaptive_stream} (a) illustrates the different elements of this block.

\paragraph{Class Array} \label{class_arr}
We established a \emph{c}-dimensional vector, named Class Array, $\mathbf{\hat{C}}$ with each dimension purposefully aligning with one of the \emph{c} 
distinct object classes recognized by the YOLOv7 algorithm. We initialize this array as a zero vector. This vector served as a flexible representation of the objects within each frame.

For instance, in a given frame, if certain objects such as `person,' `TV,' `keyboard,' `handbag,' `cell phone,' `mouse,' and `cup'  are identified and associated with class numbers 0, 4, 15, 45, 60, 78, and 79 respectively, our vector, called $\mathbf{\hat{C}}$, is dynamically updated to reflect the presence of these objects. Each time an object is found in a frame the corresponding value against its assigned index will go up by one. This process is applied to all frames in the video sequence, resulting in a comprehensive representation of the objects found within the entire video. Following the generation of the class array ($\mathbf{\hat{C}}$), we normalize it by dividing it by the total number of frames $n$. This normalization step ensures that the values in the $\mathbf{\hat{C}}$ are scaled appropriately relative to the number of frames considered. It's worth noting that for each video clip, we obtain a $\emph{c}$-dimensional array. Therefore, if our dataset comprises \emph{m} clips, we will have $\emph{m}\times\emph{c}$-dimensional arrays. Algorithm \ref{alg:array} summarize the process of creating $\mathbf{\hat{C}}$.%class array.

\begin{equation} \label{class_arr_eq}
      \mathbf{\hat{C}} = \{x_i|x \in \mathbb{R}^{1\times \emph{c}}, 1 \leq i \leq m\}   
\end{equation}%\\

\begin{algorithm}[t]
\caption{Creating Class Array ($\hat{C})$}
\label{alg:array}
%\begin{algpseudocode}
\textbf{Input:} Input Frames

\textbf{Output:} Class array ($\hat{C}$)

\textbf{Step 1:} Define a zero vector, $\hat{C}$ of shape (1,\emph{c}).

\textbf{Step 2:} Loop through all the frames in the stack.

\textbf{Step 3:} For each object present in the frame add the class of that object to the value of corresponding index.

$\hat{C}[i] +=$ \emph{number of that object found in current frame} , $i =$ \emph{index corresponding to that object.}

\textbf{Step 4:} Normalize the array by dividing it with the number of frames $n$.

$\hat{C} /= n$, $n$ = number of frames in the video.

\textbf{Return:} $\hat{C}$
%\end{algpseudocode}
\end{algorithm} 

%\textbf{\emph{Learnable Decision Embedding:}}\label{LDE}
\paragraph{Learnable Decision Embedding} \label{LDE}
The learnable decision embedding is a set of parameters that determines whether to use the raw or masked frame stack for forward propagation. Represented by weight parameters $\mathbf{W} \in \mathbb{R}^{c \times 1}$, this module functions like trainable weights, adjusted during back-propagation. It enables the model to adaptively refine its decision-making, improving encoder independence and overall performance as the model learns to make informed choices based on the input data.

%\textbf{\emph{Multilayer Perceptron (MLP) Network:}} \label{mlp}
\paragraph{Multilayer Perceptron (MLP) Network} \label{mlp}

This neural network architecture includes a single hidden layer designed to process input arrays or tensors of size $\mathbb{R}^{1 \times c}$, where $c$ represents the number of classes (80 in our experiments). The hidden layer outputs a scalar value with a Sigmoid activation. This network aids in two key decisions: selecting the appropriate input frame stack (masked or raw) for forward propagation and determining which frames should receive more attention. The class array (\ref{class_arr}) and frame tensor (\ref{frame-tensor}) serve as inputs to this MLP network. The MLP module outputs:

\begin{equation} \label{mlp_output1}
       D_1 = \sigma(\xi(\mathbf{\hat{C}}))
\end{equation}
\begin{equation} \label{mlp_output2}
       \mathbf{A} = \sigma(\xi(\mathbf{\hat{F}}))
\end{equation}

where, $\xi$ denotes the MLP network, $\sigma$ is the Sigmoid activation function, $\mathbf{\hat{C}}$ is the class array, and $\mathbf{\hat{F}}$ is the frame tensor. $D_1$ helps decide which frame stack to use, while $\mathbf{A}$ is an attention vector with $\mathbf{A} \in \mathbb{R}^{n \times 1}$, where $n$ is the number of frames in the input stack.

%\textbf{\emph{Matrix Multiplication Block:}} \label{mat_mul}
\paragraph{Matrix Multiplication Block} \label{mat_mul}

This block is responsible for conducting the conventional matrix multiplication operation. It takes two inputs: the $\mathbf{\hat{C}}$ described in Eqn. \ref{class_arr_eq}, 
and the learnable decision embedding vector $\mathbf{W}$  (\ref{LDE}). These two matrices are multiplied together to yield a scalar value, which can assume any real number. The multiplication process is expressed in following equation. 

\begin{equation} \label{mat_mul_eq}
    \begin{bmatrix}
        C_1 & C_2 & \ldots & C_{\zeta}
    \end{bmatrix}
    \begin{bmatrix}
        w_1 \\
        w_2 \\
        \vdots \\
        w_{\zeta}
    \end{bmatrix}
    = D_2
\end{equation}

Here, $\zeta$ denotes the number of object returned by YOLOv7 algorithm. In our experiment, we set this value to $80$. The scalar value $D_2$ from Eqn. \ref{mat_mul_eq} is subtracted from $D_1$ (Eqn. \ref{mlp_output1}) to yield $D$ (Eqn. \ref{cal_d}). This result is then processed through a Rectified Linear Unit (ReLU) activation function, producing $\hat{D}$ (Eqn. \ref{final_decision}). The ReLU function determines whether to apply subsequent processing to raw or masked frame data based on the subtraction result.

\begin{equation} \label{cal_d}
    D = D_1 - D_2
\end{equation}
\begin{equation} \label{final_decision}
    \hat{D} = \text{ReLU}(D)
\end{equation}%\\

%------------------------------------------------------------------start from here----------------------------------------------------------------------
\paragraph{Decision Making}

As discussed, \(\hat{D}\) (Eqn. \ref{final_decision}) determines whether to use the raw or masked frame stack in the feature encoder. The ReLU function, which outputs zero for non-positive inputs and returns positive inputs unchanged, ensures that \(\hat{D}\) can only be either positive or zero. A positive \(\hat{D}\) selects the raw frame stack, while a zero value selects the masked frame stack.

Our reasoning behind this choice comes from the idea that when there are lot of objects in the frame then focusing on the objects is more important for classification than focusing anywhere else, therefore mask frames stack becomes the best choice for input frame stack. When there are lot of object in the video frames then $\mathbf{\hat{C}}$ mostly have non-zero values and therefore $D_1$ and $D_2$ becomes larger. However, due to $\sigma$ activation $D_1$ value remains in the range of $ 0 \leq D_1 \leq 1 $ whereas $D_2$ can be any real value. So, Eqn. \ref{final_decision} always returns a non-positive value in this case, which consequently lead to selecting mask frames stack for forward propagation. If \(D_1\) were not constrained to the range of 0 to 1, there could be instances where \(D\) becomes positive, resulting in the selection of the raw frame stack when the masked frame stack should have been chosen as the input.

\begin{equation} 
\mathbf{X} = 
\begin{cases}
  \mathbf{M} & \text{if $\hat{D}=0$} \\
  \mathbf{R} & \text{otherwise}
\end{cases}
\end{equation}

Here, $\mathbf{X}$ represents the frame stack vector that is forwarded into feature encoder. $\mathbf{M}$ represents mask frame stack whereas $\mathbf{R}$ represents the original raw frame stack.

\subsubsection{Framewise Attention Mechanism}

All $n$ frames extracted and sampled from video data are not equally important as not all of them will contain relevant information. Therefore, it's inefficient for a model to treat all frames equally. To address this, we have introduced the framewise attention mechanism. The output of this block is the vector \(\mathbf{A}\), which contains attention values assigned to each frame. In essence, this block enables the feature encoder to understand the significance of individual input frames for activity classification by assigning appropriate attention weights to each frame. Fig. \ref{adaptive_stream} (b) illustrates the various components of this mechanism, which will be discussed in detail below.%\\

%\textbf{\emph{Frame Tensor:}} \label{frame-tensor}
\paragraph{Frame Tensor} \label{frame-tensor}
The first step in implementing this mechanism is to create a frame tensor, $\mathbf{\hat{F}}$ which is similar to the $\mathbf{\hat{C}}$ discussed in \ref{class_arr}.

When we created $\mathbf{\hat{C}}$, we began with a single array and updated it to account for all objects detected in the frame stack. 
However, the key difference between $\mathbf{\hat{F}}$ and  $\mathbf{\hat{C}}$ is that instead of summing object numbers across frames, we generate a distinct $(1,\emph{c})$ dimensional array for each 
frame and stacked the $n$ vectors corresponding to $n$ sampled frames in the video segment to generate $\mathbf{\hat{F}}$. If there are $m$ video clips in the dataset then $\mathbf{\hat{F}}$ can be defined as follows:

\begin{equation} \label{frame_tensor_eq}
      \mathbf{\hat{F}} = \{x_i|x \in \mathbb{R}^{n\times \emph{c}}, 1 \leq i \leq m\}
\end{equation}

Let's consider the previous example: within a single frame, objects like `person,' `TV,' `keyboard,' `handbag,' `cell phone,' `mouse,' and `cup' were identified and associated with class numbers 0, 4, 15, 45, 60, 78, and 79 respectively. Previously, we dynamically updated a vector called $\mathbf{\hat{C}}$ to reflect the presence of these objects. Now, instead of updating a single array, we create $(1,\emph{c})$ dimensional arrays for each of the $n$ frames in the video clip and stack them together. Thus, our resulting $\mathbf{\hat{F}}$ is a $(n,c)$ dimensional matrix, where $n$ corresponds to the number of frames within the frame stack. Algorithm \ref{alg:tensor} summarize the procedure of creating $\mathbf{\hat{F}}$. %\\

\begin{algorithm}[t]
\caption{Calculating Frame Tensor $\mathbf{\hat{(F)}}$}
\label{alg:tensor}
%\begin{algpseudocode}
\textbf{Input:} Input Frames

\textbf{Output:} Frame Tensor ($\mathbf{\hat{F}}$ )

\textbf{Step 1:} Define $n$ number of $(1,\emph{c})$ dimensional zero vector, $\hat{f}_j$, $j$ = $j^{th}$ frame in the input frame stack, $j = 1, 2,\ldots , n.$

\textbf{Step 2:} Loop through all the frames in the stack.

\textbf{Step 3:} For each frame, count the number of every object present in the frame.

\textbf{Step 4:} Update the components of corresponding array of the frame.

$\hat{f}_j[i] +=$ \emph{number of a particular object found in $j^{th}$ frame}, $i =$ index.

\textbf{Step 5:} Stack all the arrays together and form a tensor.

$\mathbf{\hat{F}}$  = stack($\begin{bmatrix}f_1 & f_2 & \ldots & f_{n}\end{bmatrix}$).

\textbf{Return:} $\mathbf{\hat{F}}$ 

%\end{algpseudocode}
\end{algorithm}

%\textbf{\emph{Framewise Attention Vector:}}
\paragraph{Framewise Attention Vector}

In the previous section, we discussed the creation of \(\mathbf{\hat{F}}\), which this block processes to enhance the encoder's focus on relevant frame information. From Eqn. \ref{mlp_output2} we obtain a vector of attention weights defined as $\mathbf{A}\in  \mathbb{R}^{n\times 1}$ where $n$ denotes the number of frames in the input frame stack. Since, the values come from a $\sigma$ activation therefore it remains in the range of $ 0 \leq a_1 \leq 1$ where $a_1 \in \mathbf{A}$. 

%This processed vector serves as a set of weightage coefficients. These coefficients are multiplied to the corresponding frames within the input stack before they are passed through the feature encoder. This step allows the model to assign varying levels of importance to each frame, focusing more on frames that are deemed crucial for classifying activity recognition. 
This processed vector acts as a set of weight coefficients, multiplies to each frame in the input stack before passing through the feature encoder. This enables the model to prioritize crucial frames for more accurate activity recognition. 
Finally, the selected frames undergo feature extraction through a pre-trained feature encoder specialized in activity classification. Detailed information about the feature encoder is provided in Sec. \ref{exp-result}.

\section{Experiments and Results}\label{exp-result}

In this section, we first briefly discuss about the BON Dataset \cite{tadesse2021bon} used in all of our experiments. Then we discuss the experimental details along with the results. We also study the effectiveness of each component of our method from the systematic ablation experiments described here. 

\subsection{Dataset}

\begin{figure}[t]
    \centering
    \includegraphics[width = \linewidth]{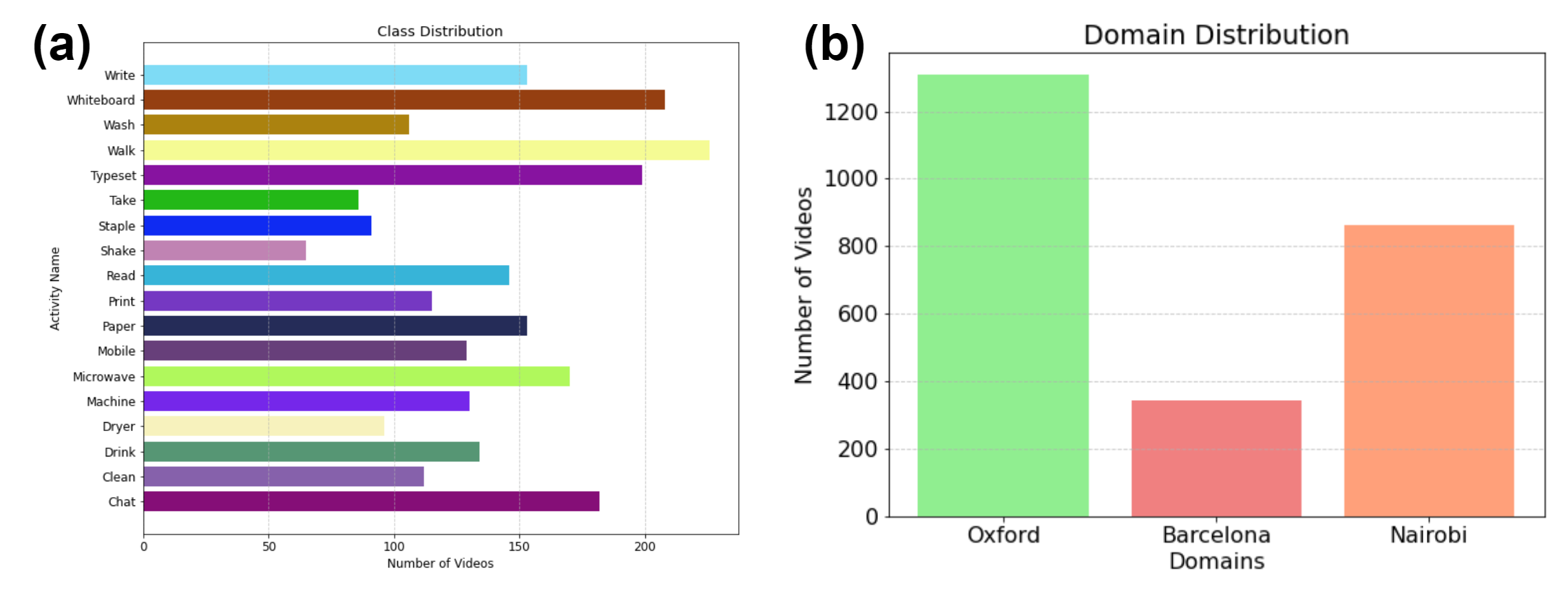} 
    \caption{(a) Class distribution and (b) Domain distribution in BON dataset. \cite{tadesse2021bon}.}
    \label{dataset}
\end{figure}

%The dataset used in this work has been extended from \cite{abebe2019first} and includes FPVs collected in different office settings and geographical locations (Barcelona, Oxford and Nairobi) which is called BON Egocentric vision dataset \cite{tadesse2021bon}. The videos in this dataset were collected using a chest-mounted GoPro Hero3+ Camera with a resolution of 1280 × 760 pixels and a 30 fps frame rate. All the videos are captured by a total of 25 subjects in the three geographical locations. There are four types of human activities present in the BON dataset: (i) ambulatory motion, (ii) human to human interaction, (iii) human to object interaction, and (iv) solo activity. Overall, the dataset contains a total of 18 activity classes. 
The dataset used in this work, the \textbf{BON Egocentric Vision Dataset} \cite{tadesse2021bon}, has been extended from \cite{abebe2019first} and includes first-person videos (FPVs) collected from various office settings and geographical locations (Barcelona, Oxford, and Nairobi). The dataset features videos recorded with a chest-mounted GoPro Hero3+ Camera at a resolution of 1280 × 760 pixels and a frame rate of 30 fps, captured by 25 subjects across these locations. It covers four types of human activities: (i) ambulatory motion, (ii) human-to-human interaction, (iii) human-to-object interaction, and (iv) solo activity, totaling 18 activity classes. Figure \ref{dataset} shows the class and domain distribution of the dataset.

As evident in Fig. \ref{dataset} (a), there is a significant class imbalance exists in the dataset, which makes it more challenging to analyze. Furthermore, the notable differences in office environments between the locations make it particularly suitable for studying domain generalization. Offices in Barcelona and Oxford are similar, despite being sourced from developed nations \cite{countryClassification}, while the office environment in Nairobi, located in a developing nation, introduces substantial environmental and demographic variations \cite{countryClassification}. These variations, including differences in skin color and environmental conditions, create a significant domain shift. The data distribution across different domains is visualized in Fig. \ref{dataset} (b).

Given these factors, this dataset presents a valuable opportunity for evaluating domain generalization. Since we want to address this domain shift problem and propose methods that enable the video classifier to generalize effectively across different domains, we designate the Oxford data for training and Barcelona data for seen domain testing as we consider it to be our source domain. And we use the Nairobi data as our target domain and use as our test data. All the results are reported for this test data. The dataset comprises 1388 video segments for training, 862 segments for target domain testing, and 246 segments for seen domain testing.

%Furthermore, the office environment in Barcelona and Oxford are quite similar as they are sourced from developed nations \cite{countryClassification}. However, there is a significant environmental change in office space for Nairobi data as it is situated in Africa, a developing nation \cite{countryClassification}. Moreover, the skin color of people involved in the Nairobi video clips are also significantly different and creates a notable domain shift in the dataset. 
%The data distribution across different domains is visualized in Fig. \ref{domain_distribution}. 
%Since we want to address this domain shift problem and propose methods so that the video classifier can generalize across domains, we designate the Oxford data for training and Barcelona data for seen domain testing as we consider it to be our source domain. And we use the Nairobi data as our target domain and use as our test data. All the results are reported for this test data. The number of video segments in training and testing were 1388 and 862, respectively. Additionally, 246 video segments were used for seen domain testing. %purposes.

\subsection{Implementation Details}
%We perform the experiments using PyTorch on a system with a Geforce RTX 2070 8 GB GPU. 
We used a batch size of 1, Adam optimizer with an initial learning rate $10^{-4}$, cross-entropy loss for the classification loss function, and accuracy and F1 Score for the evaluation metric. Additionally, we used precision and recall for ablation studies and comparing our method's performance with state-of-the-art's performance on the task. %Batch size 1 was used to mitigate the out of memory error and to compensate its effect we used gradient accumulation technique. 
Batch size of 1 was used along with gradient accumulation technique where we updated the model weights in every 32 steps.
All the experiments were run for 15 epochs. %Since, we did not do any hyper-parameter tuning, we didn't require a validation set. 
We saved the weights for all 15 epochs it was trained for and reported the best test results among the 15 epochs. During training, all the video frames were resized to 224 x 224 size and each video segments had 32 frames with a sampling rate of 3.

\subsection{Ablation Studies}

To validate the effectiveness of each proposed component, we perform ablation experiments on the BON \cite{tadesse2021bon} dataset according to the previously mentioned split. In the following sub-sections we discuss the results of all our experiments. 

\subsubsection{Feature Encoder}

 We evaluate the performance with various state-of-the-art video classification networks, namely - ResNet(2+1)D\cite{8578773}, R3D\cite{8578773}, and MC3D\cite{8578773}, CSN\cite{tran2019video}, C2D\cite{wang2018non}, I3D\cite{wang2018non} and MViT\cite{fan2021multiscale} (which are readily available from PytorchVideo library\cite{fan2021pytorchvideo}) as our backbone feature extractor to study the generalization of our proposed technique over the different encoders. Initially, we train each of the encoders with the raw frames and tested on seen domain and unseen domain's test dataset. The results are shown in Fig. \ref{chart}. From the figure it is apparent that there is a huge performance gap in between the performance on seen domain and unseen domain. Next we train the encoders for the mask frames. Finally, the encoders was trained with our proposed MLP network, decision embeddings and framewise attention mechanism incorporated into the framework and reported the results on the unseen test dataset. All these experiments show a consistent performance improvement with our proposed approach for the different encoders, as shown in  Table \ref{compared_models}. It is apparent that the MViT\cite{fan2021multiscale} network performs much better (F1 and Accuracy) when using our approach and thus, we use this encoder for the subsequent experiments.

\begin{table}[t]
\begin{threeparttable}
\caption{Performance Comparison on Different Encoders}
\label{compared_models}
\small
\begin{tabularx}{\columnwidth}{XXXXX}
\hline
\textbf{FE} & \textbf{RD Acc} & \textbf{RD F1} & \textbf{PM Acc} & \textbf{PM F1} \\ \hline
R3D\cite{8578773} & 0.4513 & 0.3460 & 0.4745 & 0.3906 \\ \hline
MC3D\cite{8578773} & 0.4640 & 0.3820 & 0.5012 & 0.4206 \\ \hline
C2D\cite{wang2018non}  & 0.5081 & 0.4509 & 0.5452 & 0.5027 \\ \hline
I3D\cite{wang2018non}  & 0.5650 & 0.5303 & 0.5974 & 0.5649 \\ \hline
CSN\cite{tran2019video}  & 0.6044 & 0.5539 & 0.6148 & 0.5732 \\ \hline
MViT\cite{fan2021multiscale} & 0.6705 & 0.6216 & 0.7146 & 0.6873 \\ \hline
\end{tabularx}

\begin{tablenotes}
      \footnotesize
      \item FE: Feature Encoder, RD: Raw Data, PM: Proposed Method.
    \end{tablenotes}
\end{threeparttable}
\end{table}

\subsubsection{Effectiveness of Decision Embedding}

\begin{figure}[t]
    \centering
    \includegraphics[width = 0.6\linewidth]{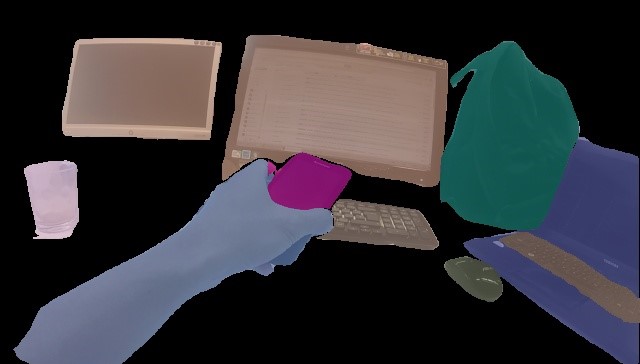} 
    \caption{Processed Mask Frame from Typeset Activity.}
    \label{processed_mask_frame}
\end{figure}

After training with the raw frames, we processed the videos frames with a YOLOv7\cite{wang2022yolov7} algorithm. We used its off-the-shelf semantic segmentation capabilities and masked the raw frames. It returns eighty different object's (COCO\cite{lin2014microsoft} object) mask if it is present in the frame. Thus after obtaining the object mask we color coded the object with a unique color and set all the remaining pixel values as zero (black). Furthermore, when color coding the objects we used 50\% of the original color of the object and 50\% the coded color. This is largely inspired by our previous work on the same dataset \cite{sushmit2020segcodenet}. An example of a processed masked frame can be seen in Fig. \ref{processed_mask_frame}. In Table \ref{ablation}, from 1st and 2nd row we see the comparison of performance change by training with raw frames and mask frames. There is a minor increase in performance. However, the model tends to fail on different activity class when trained with raw frames and masked frames subsequently. To tackle it, we introduce the decision embeddings learnable parameter. This parameter along with the object information in the stack of frames generated using YOLOv7 \cite{wang2022yolov7} helped decide the whether the forward pass for the encoder should be made with raw frame stack or masked frame stack. Incorporating this decision embedding into the framework significantly improve the feature encoder's performance as it is evident from row 3 of Table \ref{ablation}.

\subsubsection{Effectiveness of Frame-wise Attention}

\begin{figure*}[t]
    %\hspace{-20mm}
    \centering
    \includegraphics[width=\linewidth]{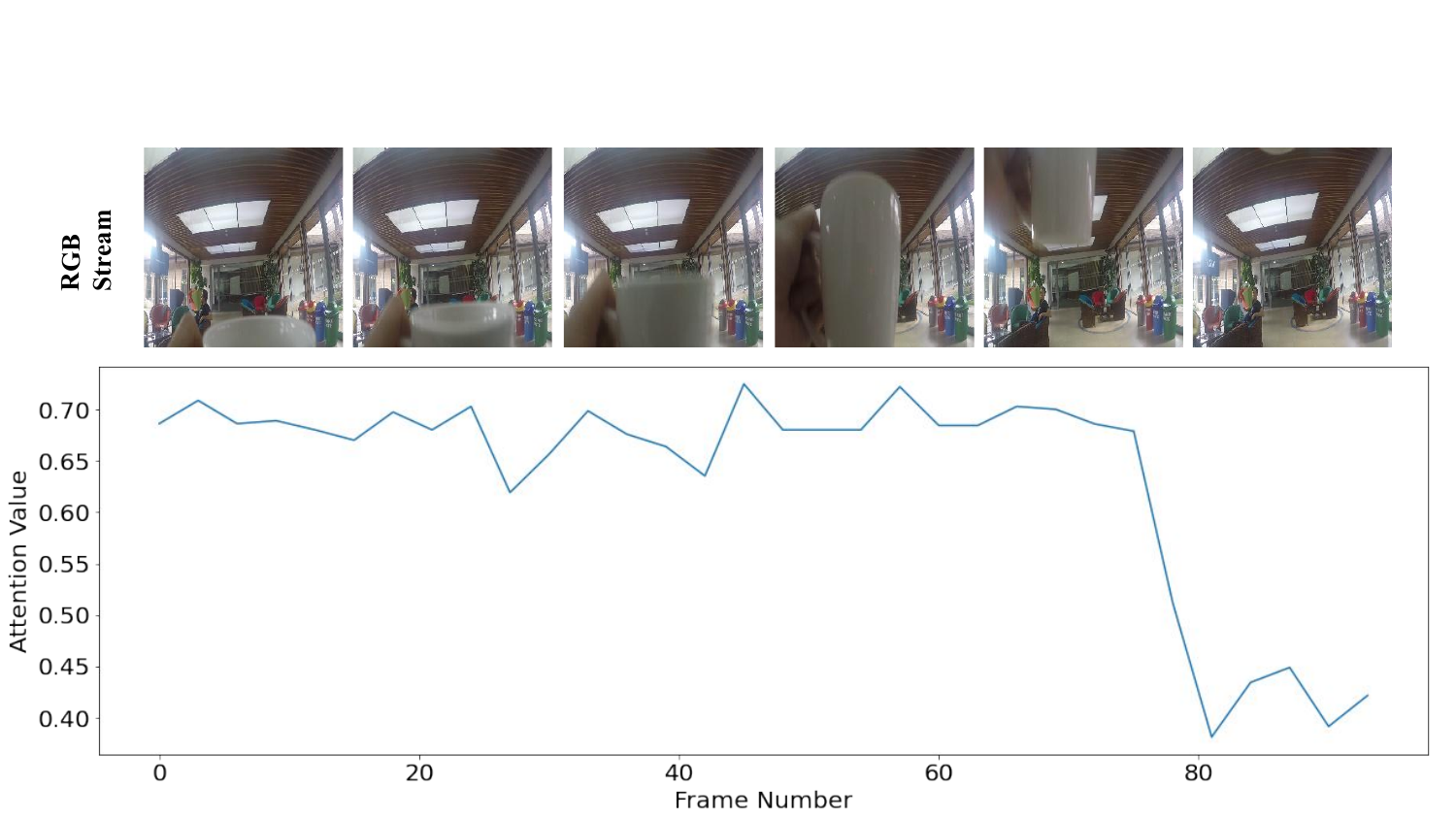}%[height=3.5in,width=7.2in]
    \vspace{-3mm}
    \caption{Variation of the frame-wise attention values for a series of sub-sampled frames obtained from a video segment containing the \emph{drink} activity class. Raw frames are shown for illustration. The frame-wise attention values is high when the person is drinking from the cup or holding the cup or the cup is visible in the frame.}
    \label{framewise-attention}
\end{figure*}

To study the effectiveness of the frame-wise attention module, we ran a set of experiments including and excluding this module, while other aspects of the framework remained the same. The results of these experiments presented in Table \ref{ablation} row 3 and 4. It shows that the frame-wise attention provides an absolute gain in all of the evaluation metrics. The mean Precision, recall, F1 score and accuracy reached $70.73\%$, $66.84\%$, $68.73\%$ and $71.46\%$, respectively. As an illustrative example, the frame-wise attention values obtained from a video sample containing the activity class \emph{chat} is shown in Fig. \ref{framewise-attention}. We observe from this figure that the attention value is higher when the cup is visible in the video frame indicating the activity ``Drink".

\begin{table}[t]
  \caption{Ablation Study (on MViT Encoder). The RD column denotes the use of the raw image data, and the MD column denotes the use of masked image data.}
  \label{ablation}
  \centering
  \begin{tabularx}{\linewidth}{XXXXXXXX}
    \toprule
    \textbf{RD} & \textbf{MD} & \textbf{DE} & \textbf{FA} & \textbf{P} & \textbf{R} & \textbf{Acc} & \textbf{F1} \\ \midrule
    $\checkmark$ & - & - & - & 0.6636 & 0.5846 & 0.6705 & 0.6216 \\
    - & $\checkmark$ & - & - & 0.6449 & 0.6336 & 0.6717 & 0.6392 \\
    $\checkmark$ & $\checkmark$ & $\checkmark$ & - & 0.6878 & 0.6193 & 0.6961 & 0.6518 \\ \midrule
    \textbf{$\checkmark$} & \textbf{$\checkmark$} & \textbf{$\checkmark$} & \textbf{$\checkmark$} & \textbf{0.7073} & \textbf{0.6684} & \textbf{0.7146} & \textbf{0.6873} \\ \bottomrule
  \end{tabularx}
  \begin{tablenotes}
    \footnotesize
    \item RD: raw image data, MD: masked image data, DE: decision embeddings, FA: framewise attention, P: Precision, R: Recall.
  \end{tablenotes}
\end{table}

% \begin{table}[t]
% \caption{Ablation Study (on MViT Encoder). The RD column denotes the use of the raw image data and the MD column denotes the use of masked image data.}
% \label{ablation}
% \centering
% \resizebox{3.4in}{0.8in}{
% \begin{tabular}{cccccc}
% \toprule
% RD & MD & DE & FA & Acc & F1 \\ \midrule
% \checkmark & - & - & - & 0.6705 & 0.6216 \\
%  - & \checkmark & - & - & 0.6717 & 0.6392 \\
% \checkmark & \checkmark & \checkmark & - & 0.6961 & 0.6518 \\
% \midrule
% \pmb{\checkmark} & \pmb{\checkmark} & \pmb{\checkmark} & \pmb{\checkmark}  & \textbf{0.7146} & \textbf{0.6873} \\
% \bottomrule
% \end{tabular}
% }
% \begin{tablenotes}
%       \footnotesize
%       \item RD: raw image data, MD: masked image data, DE: decision embeddings, FA: framewise attention.
%     \end{tablenotes}
% \addtolength{\tabcolsep}{10pt}
% \end{table}

\subsection{Comparison with SoTA method}

Domain generalizability in human activity recognition hasn't been addressed in the literature specially in office activity recognition task. Most of the works that talks about domain generalizability are often used only image data to showcase the robustness of their algorithm \cite{li2018learning,ganin2016domain, sagawa2019distributionally, huang2020self}. However, there are some activity recognition task which tackles the problem of domain adaptation, not generalization. In domain adaptation the algorithm is trained with unlabeled target data to better adapt with the target domain. We compared our proposed domain generalization pre-processing techniques incorporated in the current best video encoder's (MViT) result with the most recent domain adaptation-based human activity recognition approaches, ATCoN \cite{xu2022source} and CO$^{2}$A \cite{da2022dual}. We evaluated the results on both seen and unsenn domain test data  in Table \ref{sota-table}. From Table \ref{sota-table} we can see that our proposed method surpasses the best result by more than $5\%$ in accuracy, almost  $5\%$ in precision, more than $9\%$ in recall and more than $7\%$ in F1 on seen domain. In the unseen domain, the results are even better where our proposed method surpassed the best method by more than $14\%$ in accuracy, $18\%$ in precision, $15\%$ in recall and almost $17\%$ in f1 score.

\begin{table}[t]
\caption{Performance comparison of MViT Encoder with SoTA Domain Adaptation method}
\label{sota-table}
\begin{threeparttable} % Use this to keep notes aligned with the table
\small % Smaller font size for the table content
\begin{tabularx}{\textwidth}{l*{8}{X}} % Adjust column width using l for the first column and X for others
\toprule
\textbf{Method} & \textbf{SD \hspace{0.1cm}P} & \textbf{SD R} & \textbf{SD Acc} & \textbf{SD F1} & \textbf{UD P} & \textbf{UD R} & \textbf{UD Acc} & \textbf{UD F1} \\ \midrule 
ATCoN\cite{xu2022source} & 0.8287 & 0.7464 & 0.8150 & 0.7854 & 0.5315 & 0.5046 & 0.5684 & 0.5177 \\ 
CO$^{2}$A\cite{da2022dual} & 0.2979 & 0.2351 & 0.3031 & 0.2628 & 0.4469 & 0.3844 & 0.4599 & 0.4133 \\ \midrule
\textbf{Proposed} & \textbf{0.8784} & \textbf{0.8391} & \textbf{0.8705} & \textbf{0.8583} & \textbf{0.7131} & \textbf{0.6633} & \textbf{0.7146} & \textbf{0.6873} \\ \bottomrule
\end{tabularx}
\begin{tablenotes}
      \footnotesize
      \item SD: Seen Data, UD: Unseen Data, P: Precision, R: Recall
\end{tablenotes}
\end{threeparttable}
\end{table}

\section{Discussion} % \& Conclusion}

The results obtained from our study shed light on the domain generalization challenges in video activity recognition, particularly for the office environments. Our approach, outperforms state-of-the-art \cite{xu2022source, da2022dual} domain adaptation methods and also shows that the performance for off-the-shelf video encoder increases on unseen domain when they are incorporated with the proposed techniques. The substantial gains in accuracy, precision, recall and F1 score on these unseen domains underscore the competitive advantage of our method, making it a promising technique for real-world applications. Our ablation studies further validate the effectiveness of our strategies in addressing domain shifts.

Since we are not proposing a new architecture but instead introducing pre-processing techniques that can enhance the performance of any model across diverse domains, we aimed to compare our method specifically with other pre-processing techniques. However, to our knowledge, no existing methods are tailored for office video activity recognition that focus solely on pre-processing. As a baseline, we therefore used basic augmentation techniques used across all models trained on raw data (Table \ref{compared_models}). As shown, our methods give the video encoder a significant advantage over basic augmentation alone.

While our findings suggest potential applications in various fields, such as healthcare, surveillance, and robotics, it is crucial to acknowledge that our approach is specifically tailored to office environments. The technique was designed with the unique characteristics and challenges of office video data in mind, including variations in office layouts, lighting, and human activity patterns specific to professional settings. Although we theoretically believe our approach could extend to other environments, such as industrial or educational settings, we cannot guarantee its effectiveness without further investigation.

Additionally, we utilized YOLOv7 for object detection, which can identify 80 objects. While this was sufficient for the objects typically found in office spaces, it may not be adequate when extending to other domains where a broader range of objects is present. In such cases, more comprehensive semantic segmentation models may be required.

The performance improvements observed in our study, especially concerning environmental changes, may not be uniformly applicable across all scenarios. The effectiveness of domain generalization depends on various factors, including the degree of dissimilarity between the training and test domains. Our future research will focus on exploring the broader applicability of our approach, particularly in non-office settings, to assess its limitations and potential. We plan to investigate additional techniques to enhance domain generalization, such as leveraging more diverse datasets, incorporating reinforcement learning for fine-tuning, and exploring the fusion of multi-modal data sources. These future directions aim to build upon our current work and expand the scope of human activity recognition models to ensure they are robust across a wide range of environments and scenarios.

\section{Conclusion}

In conclusion, our study presents a novel pre-processing approach to video activity recognition that focuses on domain generalization. Our findings underline the practical significance of domain generalization techniques in the evolving landscape of video activity recognition. The success of our approach suggests promising applications across domains where recognizing activities from diverse video sources is essential. However, we acknowledge that there is room for further improvement and exploration in this area. Our study contributes to a growing body of research in HAR, offering insights that can guide future investigations aimed at achieving even more robust domain generalization.

\section*{Declaration of generative AI and AI-assisted technologies in the writing process}
During the preparation of this work a Large Language Model (ChatGPT) in order to improve language and readability of the article. 

\bibliography{mybibfile2}

\end{document}